%% file: root.tex
\documentclass[letterpaper, 10 pt, conference]{ieeeconf}  %

\IEEEoverridecommandlockouts                              %

\usepackage{graphics} %
\usepackage{epsfig} %
\usepackage{times} %
\usepackage{amsmath} %
\usepackage{xcolor}

\usepackage{mathrsfs}
\usepackage{amssymb}
\usepackage{cite}
\usepackage{dsfont}
\usepackage{xcolor}
\usepackage{mathtools}
\usepackage{mathrsfs}
\usepackage{subfig}
\usepackage{balance}
\usepackage{tcolorbox}
\usepackage{bbm}

\usepackage[english]{babel}

\usepackage{amsthm}
\newtheorem*{remark}{Remark}
\usepackage{algorithm}    
\usepackage{graphicx}
\usepackage{booktabs}
\usepackage{multirow}   
\usepackage{makecell}
\usepackage{balance}
\usepackage{wrapfig}
\usepackage[noend]{algorithmic}
\makeatletter
\apptocmd{\@maketitle}{\input{overview_fig}}{}{}
\makeatother
\title{\LARGE \bf
Active Reachability Learning for Humanoid Recoverability Monitoring
}

\author{Zeyuan Feng$^{1,2,*}$, Baljeet Singh$^{2}$, Sarthak Ranjeet Kaingade$^{2}$, Somil Bansal$^{1}$, and Bryan Trinh$^{2}$\thanks{$^{*}$ Work done while interning at Amazon.}\thanks{Correspondence: {\tt\small zeyuanf@stanford.edu}, {\tt\small somil@stanford.edu}}\\[0.3em]
{$^{1}$Stanford University \quad $^{2}$Amazon}}

\begin{document}

\maketitle
\thispagestyle{empty}
\pagestyle{empty}

\begin{abstract}
Deploying humanoid task policies requires reliable runtime monitors that anticipate unrecoverable falls in time to trigger a fallback.
We formulate this task as learning the stochastic, finite-horizon recoverability envelope of a given fallback policy. Synthesizing this monitor is challenging: existing methods assume training data naturally covers the recoverability envelope, yet acquiring such data remains a major bottleneck for contact-rich systems. Consequently, they recover only a fraction of the true envelope, leaving them prone to optimistic hallucinations for out-of-distribution (OOD) inputs. We resolve this via a unified framework that iterates between value learning and active exploration. Data exploration is reframed as a reach-avoid problem, where an uncertainty-guided explorer systematically uncovers unseen states while avoiding confidently unrecoverable regions. This loop is further supported by a combined Temporal Difference and Monte Carlo objective for accurate in-distribution value prediction and reliable uncertainty estimation. By explicitly treating the remaining high-uncertainty regions as unrecoverable, the monitor mitigates optimistic OOD hallucinations. Evaluated on stochastic benchmarks and a 29-DoF Unitree G1 humanoid, our approach drastically improves monitor coverage and accuracy and successfully transfers zero-shot to real-world hardware.
\end{abstract}

\section{Introduction}
\label{sec:intro}
\input{intro}

\section{Problem Formulation}
\label{sec:formulation}
\input{formulation}

\section{Method}
\label{sec:method}
\input{method}

\section{Results}
\label{sec:results}
\input{results}

\section{Conclusion and Limitations}
\input{conclusion}
\label{sec:conclusion}

\section*{Acknowledgments}
We thank Nicholas Morozovsky, Yinai Fan, and Shravan Tangudu for proposing the humanoid test cases.
\bibliographystyle{IEEEtran}
\bibliography{references}
\input{appendix}

\end{document}

%% file: overview_fig.tex
\centering
\label{fig:cover}
\includegraphics[width=1.0\textwidth]{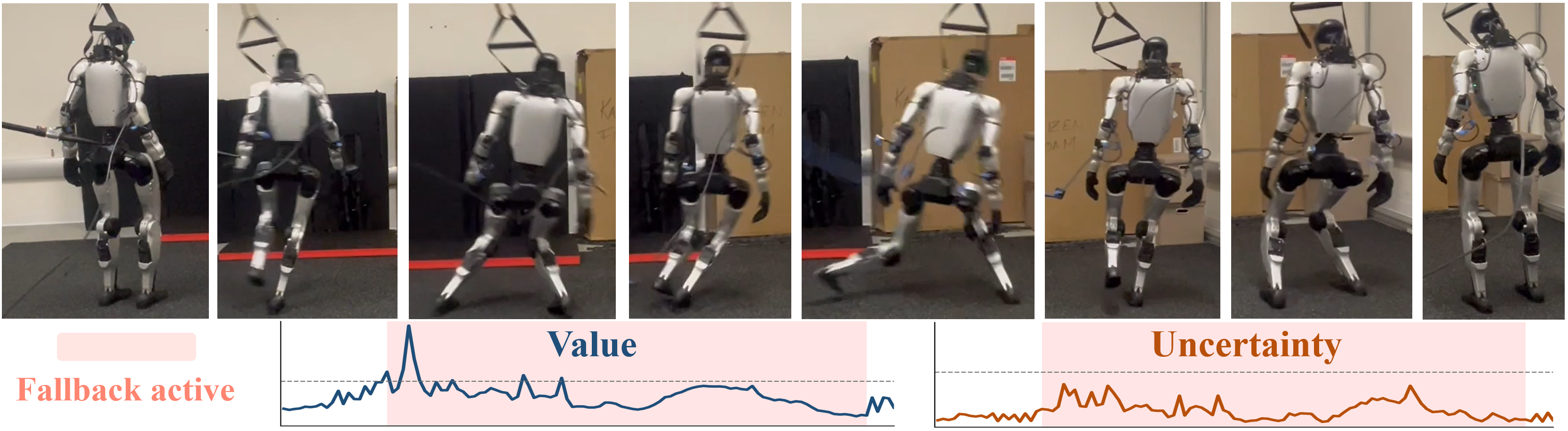}
\vspace{-0.3cm}
\captionof{figure}{Runtime recoverability monitoring using a liveness value function and uncertainty estimator on a G1 humanoid.
}
\vspace{-1em}
\setcounter{figure}{1}

%% file: intro.tex
The deployment of contact-rich robotic systems, such as humanoids, holds immense potential for automating tasks in unstructured environments. However, inherent instability and underactuation make humanoids prone to catastrophic failures, such as unrecoverable falls. Hence, safely deploying versatile task policies necessitates reliable runtime monitors that detect impending failures in advance, allowing the system to trigger a robust fallback policy before a fall becomes unavoidable. In this paper, we formulate the monitor synthesis as learning an empirical, finite-horizon recoverability (or liveness) envelope for a fixed fallback balancing policy. Such reliable monitors enable rapid development and safe deployment of \emph{arbitrary} humanoid task policies while drastically reducing the risk of hardware damage.

The synthesis of this recoverability monitor exhibits extreme numerical stiffness, as the underlying liveness value may change violently during exact rigid-body interactions. For example, a slipping foot and a firm step may share nearly identical states at the moment of contact, yet lead to completely different outcomes. More fundamentally, training requires a comprehensive mapping of the underlying liveness set, encompassing states that are recoverable and reachable under realistic physical disturbances. To cover these states, we cannot uniformly sample the state space or artificially ``teleport'' the robot to expand coverage, as minor kinematic offsets often cause rigid-body penetration, triggering severe numerical artifacts in the simulator that corrupt the training data. Therefore, we are restricted to acquiring data via episodic rollouts driven by dynamic controls or physical perturbations. However, rolling out from a manually designed initial-state distribution yields insufficient liveness set coverage, leaving the monitor prone to making overly optimistic predictions for out-of-distribution (OOD) inputs.

Existing safety frameworks struggle with the compounded dimensionality and data coverage challenges. Analytical Control Barrier Functions (CBFs)~\cite{ames2019control, yang2025cbf, bena2025geometry, yang2025shield} are effective for task-space constraints but scale poorly to whole-body recoverability monitoring. Similarly, existing stochastic reachability solvers~\cite{esfahani2016stochastic, hoshino2026physics} are computationally restricted to low-dimensional systems. Physics-Informed Neural Network (PINN) reachability methods~\cite{bansal2020deepreachdeeplearningapproach, feng2025bridgingmodelpredictivecontrol, 11127528} handle high dimensionality with deep learning, but rely on uniform state-space sampling. This renders them incompatible with contact-rich domains where uniform sampling inevitably yields mostly irrelevant or non-physical, corrupted samples.

On the other hand, Reinforcement Learning (RL)-based methods learn from simulator rollouts, which respect physical constraints. RL-based CBFs~\cite{so2024solvingminimumcostreachavoid, so2024train, morton2026constrainedwholebodytrackinghumanoid} and RL-based reachability approaches~\cite{8794107, Hsu__2021, wang2026offlinepolicyevaluationmanipulation, chen2026lambdareachabilitygeometrichorizonsafetybellman, long2026humanoidsafestoplearned} adopt this strategy. However, because these rollouts originate from nominal initial-state distributions, they only recover a narrow fraction of the true recoverability envelope. This limited coverage not only reduces the allowable operation space but also causes overly optimistic predictions for OOD states. While recent work attempts to mitigate this by treating high-uncertainty OOD states as unrecoverable~\cite{seo2025uncertaintyawarelatentsafetyfilters}, this assumes the training dataset naturally captured enough ``OOD'' states (determined by a frozen world model) prior to learning. Standard RL exploration techniques~\cite{osband2016deep, osband2019deep, tang2017exploration} also cannot fill this gap, as they are designed to maximize task reward rather than systematically chart the boundaries of recovery failure. Consequently, despite its critical importance for real-world deployment, the problem of actively exploring the recoverability envelope under non-teleporting physical constraints remains under-explored by the community.

We tackle this challenge through a unified bi-level synthesis framework. In the \textit{Inner Level (Value Learning)}, we estimate the finite-horizon liveness value of the fallback policy using a combined Temporal Difference (TD) and Monte Carlo (MC) objective, providing stable gradients for accurate in-distribution (in-D) value learning and reliable epistemic uncertainty estimation. In the \textit{Outer Level (Exploration)}, we reframe data augmentation as a reach-avoid problem, where an active explorer targets OOD states to expand coverage without appending confidently unrecoverable samples. By systematically uncovering the recoverability envelope and treating remaining high-uncertainty regions as failures, the monitor effectively mitigates optimistic hallucinations without sacrificing allowable operation space. Finally, we validate this framework on stochastic benchmarks and demonstrate zero-shot transfer on a 29-DoF Unitree G1 humanoid.

%% file: formulation.tex
\subsection{System Dynamics}
We consider a system operating within a discrete-time Markov Decision Process under a given fallback controller $\pi$. The dynamics form a stochastic system $x_{k+1} \sim \mathcal{P}(\cdot \mid x_k, a_k, d_k)$, where $x_k \in \mathcal{X}$ is the continuous state at step $k$, $a_k \in \mathcal{A}$ is the control action, and $d_k \in \mathcal{D}$ is an \emph{optional} artificial disturbance. 
Crucially, we model the system as inherently stochastic rather than relying on traditional worst-case bounded disturbances for liveness synthesis, because compounding worst-case disturbances across multiple sources can easily overpower a humanoid's control authority, yielding impractically small recoverability envelopes.

During data collection (episodic rollouts), the system is closed-loop under $\pi$, with $a_k = \pi(x_k)$ and $d_k \equiv 0$. During data exploration (novel initial state discovery), an active explorer instead freely selects the control $a_k$ and may additionally apply $d_k \neq 0$ to steer the system toward more informative states.
We assume a narrow, manually defined distribution of initial states $\rho_0(x)$ to seed the first episodic rollouts. As training progresses, subsequent rollouts can optionally be initialized from states accumulated in the replay buffer to continuously expand data coverage.

\subsection{Liveness Value Function and Safe-Set Coverage}
The controller aims to stabilize the system toward a target set $\mathcal{G} = \{x: g(x)\leq 0\}$ (e.g., set of upright stationary states). We quantify the liveness of $x$ given $\pi$ through a value function $V(x)$, defined as the expected minimum target function over all possible future trajectories:
\begin{equation}
\resizebox{0.95\linewidth}{!}{%
    $\displaystyle V(x) = \mathbb{E}_{\mathcal{P}}\!\left[ \min_{0 \le \kappa < \infty} g(x_\kappa) \,\middle|\, x_0 = x,\ a_\kappa = \pi(x_\kappa),\ d_\kappa \equiv 0 \right].$%
}
\end{equation}
A smaller liveness value $V(x)$ indicates a more recoverable current state. In the special case of a binary indicator $g(x) \in \{0, 1\}$---$0$ inside $\mathcal{G}$ and $1$ otherwise---$V(x)$ reduces to the exact probability that the system fails to reach $\mathcal{G}$.

For effective real-world monitoring, we evaluate the learned value function on two critical axes: predictive accuracy and spatial coverage. Specifically, our goal is to learn a neural value function $V_\theta(x)$ as the recoverability monitor by minimizing the approximation error over the $\epsilon$-liveness set, defined as $\mathcal{L}_\epsilon=\{x \in \mathcal{X} \mid V(x) \leq \epsilon\}$. Mathematically, we aim to minimize:
\begin{equation}
    \mathcal{J}(\theta) = \mathbb{E}_{x \sim \text{Uniform}(\mathcal{L}_\epsilon)} \left[ \left\| V(x) - V_\theta(x) \right\|^2 \right].
\end{equation}
Intuitively, coverage is unimportant in the deeply unrecoverable region where $V(x) > \epsilon$, because by mapping $\mathcal{L}_\epsilon$ accurately, the monitor can reliably alert before the system state drifts OOD. However, for underactuated systems, fully covering $\mathcal{L}_\epsilon$ using only episodic rollouts for exploration is often infeasible. Consequently, the monitor must empirically remain conservative on OOD inputs, treating unexplored regions as inherently unrecoverable to prevent hallucinations.

\subsection{Humanoid Balancing Instantiation}
We instantiate the framework on humanoid applications, where $\pi$ is a balancing policy and the stochasticity in $\mathcal{P}$ arises from observation noise and domain randomization (e.g., ground friction, body mass, and PD-gain perturbations). 
$\mathcal{G}$ denotes the set of stationary states, in which the robot's torso and joint velocity magnitudes stay within designated bounds and the torso 
height remains close to its canonical value.
We assume that if $\pi$ can recover balance from $x$, the running minimum of $g$ along the trajectory is attained within a horizon of $K$ steps.
Under this assumption, the infinite-horizon value coincides with the fixed-horizon one $V(x) \equiv V(x, 0)$, where
\begin{equation}
\label{eqn:value_def}
\resizebox{0.95\linewidth}{!}{%
    $\displaystyle V(x, k) = \mathbb{E}_{\mathcal{P}}\!\left[ \min_{k \le \kappa \le K} g(x_\kappa) \,\middle|\, x_k = x,\ a_\kappa = \pi(x_\kappa),\ d_\kappa \equiv 0 \right].$%
}
\end{equation}
Although the monitor only queries $V(x) \equiv V(x, 0)$, learning $V(x, k)$ provides a stronger signal from boundary anchoring ($V(x, K) \equiv g(x)$) and a temporal curriculum.
Concretely, we use $K = 150$ steps with $\Delta t = 0.02$\,s, i.e., we certify whether the robot returns to a stationary balanced pose within $3$\,s under $\pi$.
Assuming the fallback policy is empirically more robust than arbitrary task policies, the learned monitor can function as a universal safety layer during deployment.
\begin{remark}[Practical Consideration: Reach-Avoid Extension]
The liveness formulation assumes that reaching $\mathcal{G}$ means $\pi$ has recovered without collapsing to the ground and allows the system to stay stationary forever. However, highly agile recovery controllers may allow the system to temporarily collapse then recover, and deployment may impose strict collision-avoidance constraints to prevent damage to the surrounding environment and humans. For these scenarios, our framework naturally extends to reach-avoid formulations.
\end{remark}

%% file: method.tex
\subsection{Overview and Active Learning Loop}
\label{subsec:overview}
Because manually defining the edge of the recoverability envelope for a high-dimensional, contact-rich humanoid is intractable, we cast the monitor synthesis as an iterative active-learning loop. The framework alternates between accurately learning the liveness value of known states and actively discovering novel states at the fringe of the policy's capabilities. We begin by rolling out the given policy $\pi$ from $\rho_0(x)$ to collect an initial dataset $\mathcal{D}$. Using this dataset, we learn an accurate in-D value function alongside an uncertainty estimator. 
Leveraging this uncertainty as a reward signal, and initializing from states already in $\mathcal{D}$, we then train a disturbance explorer to actively push the system toward novel OOD states that are at most marginally unrecoverable. 
If the explorer discovers a sufficient number of high-uncertainty states, we roll out the policy $\pi$ from these newly discovered states to augment $\mathcal{D}$ and return to the value-learning phase. Once the explorer fails to find sufficient meaningful new states, the loop terminates, and the framework outputs the learned value function and uncertainty estimator as the recoverability monitor. For our primary humanoid application, during the deployment of an arbitrary task policy, the monitor triggers the balancing policy $\pi$ until a stationary state is reached whenever either the liveness value or uncertainty exceeds a predefined threshold ($\epsilon_V \ll \epsilon$). The full active-learning loop is shown in Alg.~\ref{alg:dagger_loop}.
\vspace{-2mm}
\begin{algorithm}[h]
\caption{Uncertainty-Guided Monitor Synthesis Loop}
\label{alg:dagger_loop}
\begin{algorithmic}[1]
\REQUIRE Policy $\pi$, initial-state distribution $\rho_0$
\STATE Initialize dataset $\mathcal{D}$ by rolling out $\pi$ from $\rho_0$.
\FOR{round $r = 1, 2, \dots, N_\text{rounds}$}
    \STATE Fit $V_\theta(x, t)$ via~(\ref{eqn:total_loss}) and uncertainty $\sigma(x)$ on $\mathcal{D}$.
    \STATE Form $g_{\text{explore}}$ and $l_{\text{explore}}$ from $V_\theta, \sigma$.
    \STATE Fine-tune $V_{\text{explore}}(x, t)$ and $\pi_{\text{explore}}$ over a medium horizon (warm-started when possible).
    \STATE Roll out $\pi_{\text{explore}}$ from $x_0 \sim \mathcal{D}$ to find novel seed states.
    \STATE Execute $\pi$ from seeds to collect data; append to $\mathcal{D}$.
    \IF{fewer than $N_{\min}$ novel seeds found}
        \STATE \textbf{break} 
    \ENDIF
\ENDFOR
\RETURN Recoverability monitor $V_\theta(\cdot)$ and $\sigma(\cdot)$
\end{algorithmic}
\end{algorithm}
\vspace{-3mm}
\subsection{In-Distribution Value Function Learning}
We learn a discrete-time value $V_\theta(x, k)$ that evaluates the expected recoverability of states given policy $\pi$ over a finite horizon
$k \in \{0, 1, \dots, K\}$. 
The value is the expected minimum target function achieved along the trajectory, as defined in (\ref{eqn:value_def}).
The induced liveness set is $\{x : V_\theta(x, 0) \leq \epsilon_V \}$.
Crucially, this induced set dictates which states the monitor \emph{treats} as recoverable at deployment ($\epsilon_V \ll \epsilon$), distinct from the broader $\mathcal{L}_\epsilon$ mapped during synthesis.

Rather than a single-step TD backup, we train $V_\theta$ using multi-step (TD-$\lambda$) targets that mix an observed running minimum with a bootstrap. Specifically, for a window of $m$ steps we form the running minimum $g_{k:k+m} = \min_{\kappa=k}^{k+m-1} g(x_\kappa)$ and the TD-$\lambda$ target
\begin{equation}
    \label{eqn:td_lambda_target}
    G_\theta(x_k, k) = \min\bigl(g_{k:k+m},\; \operatorname{sg}\!\bigl[V_\theta(x_{k+m}, k+m)\bigr]\bigr),
\end{equation}
where $m$ is drawn from a truncated geometric distribution (a TD-$\lambda$ schedule with decay $\lambda$, capped at $M$ steps or the boundary $K$), and $\operatorname{sg}[\cdot]$ is the stop-gradient operator.
The TD loss is the mean-squared residual,
\begin{equation} \label{eq:TD_loss}
    \mathcal{L}_{TD} = \mathbb{E}_{x \sim \mathcal{D}}
    \bigl[\, \lVert V_\theta(x, k) - G_\theta(x, k) \rVert^2 \,\bigr].
\end{equation}
\begin{remark}[Conservative bootstrap bias in stochastic systems]
Let $Y_{k+m} := \min_{\kappa=k+m}^{K} g(x_\kappa)$, so that $\min_{\kappa=k}^{K} g(x_\kappa) = \min(g_{k:k+m}, Y_{k+m})$ and $\mathbb{E}[Y_{k+m} \mid x_{k:k+m}] = V(x_{k+m}, k+m)$. As $\min(g_{k:k+m}, \cdot)$ is concave, Jensen's inequality gives
\begin{equation*}
\begin{aligned}
    V(x, k) &= \mathbb{E}\bigl[\min_{k \le \kappa \le K} g(x_\kappa)\bigr] \\
    &= \mathbb{E}\bigl[\min(g_{k:k+m},\, Y_{k+m})\bigr] \\
    &\le \mathbb{E}\bigl[\min(g_{k:k+m},\, \mathbb{E}[Y_{k+m} \mid x_{k:k+m}])\bigr] \\
    &= \mathbb{E}\bigl[\min(g_{k:k+m},\, V(x_{k+m}, k+m))\bigr] \\
    &= \mathbb{E}[G(x_k, k)],
\end{aligned}
\end{equation*}
where $G$ denotes the target~(\ref{eqn:td_lambda_target}) evaluated at the exact bootstrap $V$. The $m$-step target therefore overestimates the value even under zero learning error, which \emph{shrinks} the liveness set. This bias arises only from the bootstrapped tail and vanishes at $m = K-k$, where $G$ reduces to the unbiased but high-variance MC estimate.
\end{remark}

To ground the network in actual liveness outcomes, we further include long-horizon MC supervision. 
The empirical liveness target for a sampled state $x$ over $N_\text{MC}$ rollouts is
\begin{equation}
    \hat{V}(x, k) = \frac{1}{N_\text{MC}} \sum_{i=1}^{N_\text{MC}} \min_{\kappa=k}^{K} g(x_{i,\kappa}),
\end{equation}
and the MC loss is likewise the prediction residual, $\mathcal{L}_{MC} = \mathbb{E}_{x \sim \mathcal{D}_{MC}} \bigl[\lVert V_\theta(x, k) - \hat{V}(x, k) \rVert^2\bigr]$. MC targets are far more expensive to generate, since each demands many long-horizon rollouts. In practice we therefore generate a small set of MC targets as unbiased anchors that stabilize training, while collecting abundant TD targets to enforce local temporal consistency of the value function. Together, each iteration minimizes a weighted combination,
\begin{equation}
    \label{eqn:total_loss}
    \mathcal{L}_{total} = \mathcal{L}_{TD} + \beta\, \mathcal{L}_{MC}.
\end{equation}

\textbf{Initial dataset and time curriculum.}
We construct the initial dataset in two stages. We first roll out $\pi$ from samples of the initial-state distribution $\rho_0$ to form the TD dataset. We then draw a second set of initial states---from both $\rho_0$ and the states visited in the TD dataset---and generate long-horizon rollouts from them to form the MC dataset; the initial dataset $\mathcal{D}$ is the union of the two.
Importantly, accurately learning the in-D value function is crucial for meaningful epistemic uncertainty estimation.
To stabilize temporal value learning, we adopt a standard time-curriculum strategy \cite{bansal2020deepreachdeeplearningapproach}, in which the sampled time step $k$ for the TD and MC targets is expanded gradually from $\{K-1\}$ to the full horizon $\{0, 1, \dots, K-1\}$, with denser sampling near the boundary $k = K-1$. Once the exploration loop begins and $\mathcal{D}$ grows round by round, we remove the curriculum and fine-tune $V_\theta$ directly over the full horizon on the expanded dataset.

\subsection{Uncertainty- and Reachability-Based Explorer}
To efficiently augment the existing dataset with novel but potentially recoverable states, we cast exploration as a Backward Reach-Avoid Tube (BRAT) problem. Starting from states in the replay buffer $\mathcal{D}$, we train an exploration policy $\pi_{\text{explore}}$ that drives the system into a target set of high epistemic uncertainty while staying out of an avoid set of confidently unrecoverable states.

Specifically, we first train an off-the-shelf epistemic uncertainty estimator $\sigma(x)$ (e.g., a value-function ensemble or Random Network Distillation (RND)) on the current dataset $\mathcal{D}$ to determine the high-uncertainty target set:
\begin{equation*}
    \mathcal{G}_{\text{explore}} = \{x : g_{\text{explore}}(x) = \epsilon_\sigma - \sigma(x) < 0\}.
\end{equation*}
The avoid set consists of states with high predicted failure and low uncertainty:
\begin{equation*}
\resizebox{0.95\linewidth}{!}{%
    $\displaystyle \mathcal{A}_{\text{explore}} = \{ x: l_{\text{explore}}(x) = \max\bigl(\epsilon - V_\theta(x, 0),\; \sigma(x) - \epsilon_\sigma \bigr) \leq 0 \},$%
}
\end{equation*}
that is, $l_{\text{explore}}(x) \leq 0$ requires \emph{both} $V_\theta(x, 0) > \epsilon$ (likely unrecoverable) \emph{and} $\sigma(x) < \epsilon_\sigma$ (low uncertainty). Importantly, the monitor triggering value $\epsilon_V$ needs to be substantially smaller than the exploration stopping value $\epsilon$ for robust deployment. Both $g_{\text{explore}}$ and $l_{\text{explore}}$ are continuous, providing a dense gradient signal even far from their boundaries. 
Consequently, even when $\mathcal{G}_{\text{explore}}$ is effectively empty, the learned $\pi_{\text{explore}}$ still moves toward relatively more novel states, and the system's momentum helps uncover genuinely new data.

We solve for the exploration value function $V_{\text{explore}}(x, t)$ and its induced policy $\pi_{\text{explore}}$ using an existing learning-based reachability toolbox \cite{feng2026gradientfreeneuralhamiltonjacobireachability}. While the toolbox traditionally approximates an optimal time-conditioned policy $\pi_{\text{explore}}(x,t)$, we instead learn a time-independent policy $\pi_{\text{explore}}(x)$ to simplify the data exploration loop at a slight cost to optimality. The training horizon for the explorer is set as a hyperparameter $T_\text{explore}$. A moderate $T_\text{explore}$ is crucial: it prevents the ineffective exploration from a myopic horizon, while ensuring seeds do not collapse onto a few distant high-uncertainty peaks, as happens with overly long horizons.

\textbf{Data augmentation.} Each round, we roll out $\pi_{\text{explore}}$ from states sampled from $\mathcal{D}$. Notably, the rollout horizon for these exploration trajectories can extend beyond the explorer's training horizon $T_\text{explore}$ to encourage exploration. Along each trajectory, we extract states that successfully enter the high-uncertainty target set to serve as novel seeds. Because multiple valid seeds may be encountered within a single continuous rollout, we cap the extraction to at most 5 seeds per trajectory to prevent deep penetration into unrecoverable OOD regions. From each extracted seed, we then execute the fallback policy $\pi$ to generate fresh TD and MC data, which are subsequently appended to $\mathcal{D}$. Value learning and exploration alternate until a round yields fewer than $N_{\min}$ novel seeds or a predefined round budget is reached.

%% file: results.tex
\begin{figure}[t]
    \centering
    \includegraphics[width=0.75\linewidth]{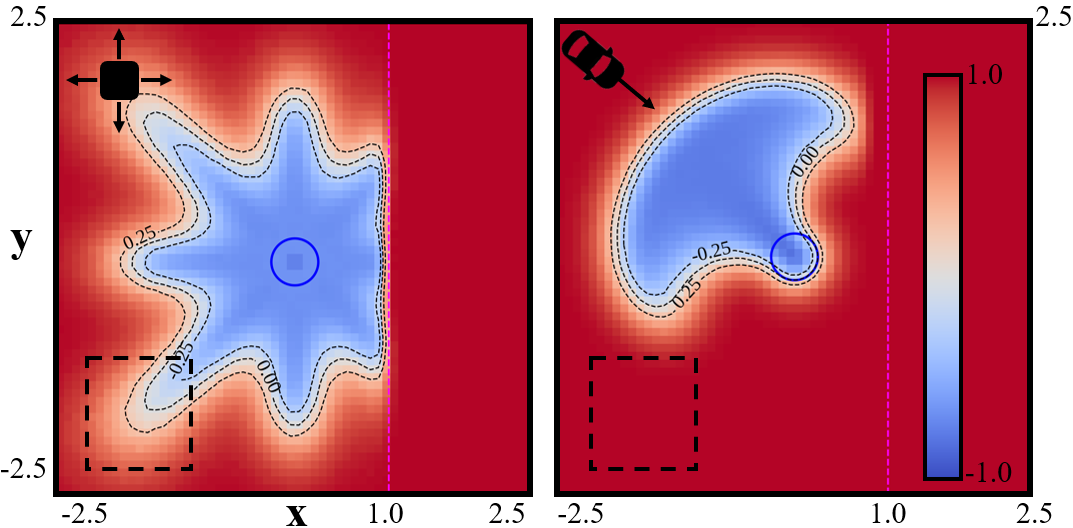}
    \caption{\footnotesize Value heatmaps and level-set contours in the $x$-$y$ plane (Left: DI; Right: Dubins). Black dashed boxes denote the initial distribution $\rho_0$ projected onto the $x$-$y$ plane (unshown dimensions are sampled uniformly). Solid circles mark the target set, defined via $g(x)=\tanh(\text{SDF}(x))$, where SDF denotes the signed distance function to the target set.}
    \label{fig:toy_systems}
    \vspace{-2em}
\end{figure}

We evaluate the framework across three settings.
First, we conduct \textbf{case studies on two underactuated benchmarks}---a stochastic 4D planar double integrator (DI) and a 3D Dubins car.
Both systems admit uniform grid sampling strictly for offline evaluation purposes, allowing us to benchmark our pipeline against near-exact ground truths and isolate individual design choices.
During monitor synthesis, only an extremely narrow initial distribution $\rho_0$ is provided for initial data collection, as illustrated in Fig.~\ref{fig:toy_systems}, and new states can only be explored by rolling out from buffered states.
Hence, the Dubins car presents severe underactuation for \emph{exploration} because rollouts are constrained by non-holonomic dynamics with a fixed velocity and bounded turn rate.
The DI, conversely, serves as a benchmark with higher stochasticity but lower underactuation.
These case studies are structured around three questions:
\textbf{(Q1)} \emph{Coverage---can our framework recover most of the recoverable states, and which design choice performs best?}
\textbf{(Q2)} \emph{Accuracy---does the TD+MC loss outperform TD($\lambda$) alone or a time-agnostic TD loss~\cite{fisac2019bridging}?}
\textbf{(Q3)} \emph{Sensitivity---how do the MC sample budget $N_{\text{MC}}$ and the explorer's training horizon affect performance?}
Second, we \textbf{scale to a Unitree G1 humanoid}, synthesizing a liveness value $V_\theta$ and failure probability $P_{\text{fail}}$ for a balancing policy on complex high-dimensional dynamics.
Third, we \textbf{close the loop} by deploying the learned monitor on a walking task policy in simulation and on hardware.

\paragraph{Metrics} We evaluate against ``ground-truth'' empirical liveness values, where each value is labeled via $1000$ MC rollouts. We report six metrics: \textbf{recall}---the fraction of truly recoverable states within the $\epsilon$-liveness set correctly identified ($\epsilon=0.5$); \textbf{precision}---the fraction of predicted-recoverable states that are genuinely recoverable, penalizing overoptimism; \textbf{$F_1$}---their harmonic mean; \textbf{MSE}---the mean squared error; \textbf{cumulative seeds}---the total exploration seeds collected across all rounds, indicating data efficiency; and confident \textbf{coverage}---the fraction of the true recoverable set covered with low epistemic uncertainty $\sigma \leq \epsilon_\sigma$ (recall restricted to in-D predictions). We also report \textbf{wall time} on an NVIDIA RTX 6000 GPU.

\paragraph{Baselines} 
\textbf{(Q1)} We compare our reach-avoid explorer against \emph{unbiased} random-action exploration and \emph{biased} random exploration that samples a constant per-episode bias before each exploration rollout. 
Each explorer is paired with either an ensemble or RND uncertainty estimator. 
All configurations share identical termination criteria: halting if a round yields $<20$ novel seeds or hits a 20-round budget.
\textbf{(Q2)} We compare our TD+MC loss against a TD($\lambda$)-only variant, similar to \cite{chen2026lambdareachabilitygeometrichorizonsafetybellman}.
\textbf{(Q3)} We sweep the MC anchor count $N_{\text{MC}}$ and explorer training horizon $T_{\text{explore}}$.

\begin{figure*}[t!]
\centering
\begin{minipage}[c]{0.65\textwidth}
  \centering
  \includegraphics[width=\linewidth]{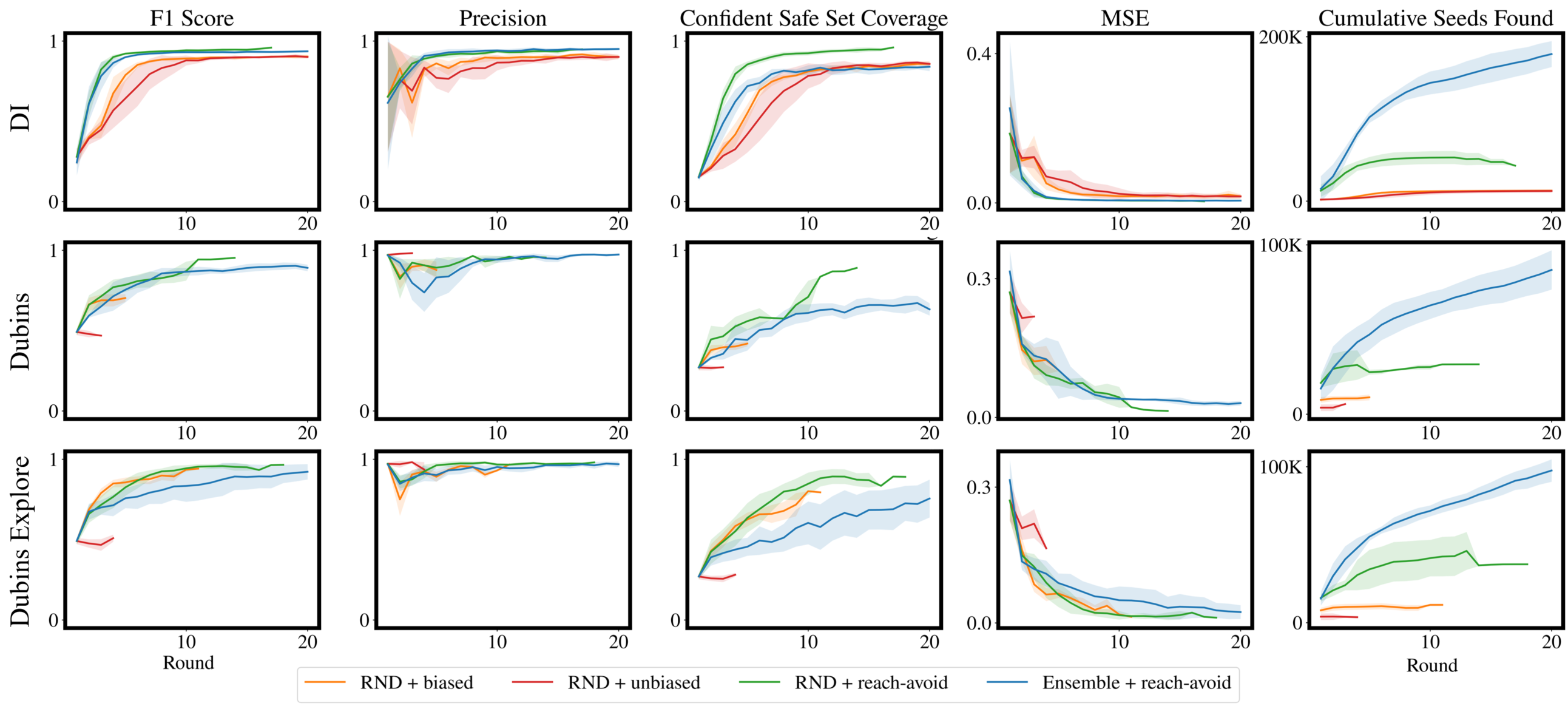}
\end{minipage}\hfill
\begin{minipage}[c]{0.33\textwidth}
  \centering
  \renewcommand{\arraystretch}{1.5}%
\resizebox{\linewidth}{!}{%
\begin{tabular}{lccccccc}
\toprule
\makecell[l]{Uncertainty /\\ Explorer} & Precision & Recall & $F_1$ & Coverage & MSE & Cum.\ seeds & Wall (min) \\
\midrule
\makecell[l]{RND + \\ unbiased} & 0.914 & 0.900 & 0.907 & 0.861 & 0.017 & 12559 & 36 \\
\makecell[l]{RND + \\ biased} & 0.919 & 0.905 & 0.912 & 0.865 & 0.017 & 12894 & 35 \\
\makecell[l]{Ensemble + \\ reach-avoid} & \textbf{0.951} & 0.922 & 0.936 & 0.840 & \textbf{0.006} & 178974 & 114 \\
\textbf{\makecell[l]{RND + \\ reach-avoid}} & 0.942 & \textbf{0.955} & \textbf{0.949} & \textbf{0.941} & \textbf{0.006} & 53524 & 71 \\
\midrule
\makecell[l]{RND + \\ unbiased} & \textbf{0.977} & 0.318 & 0.479 & 0.267 & 0.207 & 3803 & 3 \\
\makecell[l]{RND + \\ biased} & 0.902 & 0.564 & 0.690 & 0.408 & 0.118 & 9287 & 5 \\
\makecell[l]{Ensemble + \\ reach-avoid} & 0.974 & \textbf{0.821} & \textbf{0.890} & 0.632 & \textbf{0.031} & 85229 & 95 \\
\textbf{\makecell[l]{RND + \\ reach-avoid}} & 0.897 & 0.795 & 0.839 & \textbf{0.656} & 0.064 & 30254 & 33 \\
\midrule
\makecell[l]{RND + \\ unbiased} & 0.962 & 0.329 & 0.489 & 0.271 & 0.192 & 4100 & 4 \\
\makecell[l]{RND + \\ biased} & 0.931 & 0.843 & 0.884 & 0.693 & 0.043 & 10620 & 10 \\
\makecell[l]{Ensemble + \\ reach-avoid} & 0.969 & 0.885 & 0.922 & 0.756 & 0.024 & 97537 & 100 \\
\textbf{\makecell[l]{RND + \\ reach-avoid}} & \textbf{0.976} & \textbf{0.964} & \textbf{0.970} & \textbf{0.921} & \textbf{0.011} & 43035 & 60 \\
\bottomrule
\end{tabular}%
}
\end{minipage}
\caption{\footnotesize Ablation across toy systems (plot rows: DI, Dubins3D, Dubins3D-Exp.). Curves (left) show per-round mean over seeds; the table (right) reports the final-round means for the same metrics. RND uses a single value net, ensemble uses $N_{\text{ens}}$ nets. Our default is \textbf{RND + reach-avoid}.}
\label{fig:ablation}
\vspace{-1.5em}
\end{figure*}
\subsection{Case Studies: DI and 3D Dubins}
\label{subsec:res_toy}
\textbf{Systems.} The 4D DI state comprises planar position and velocity with additive Gaussian noise. The 3D Dubins car state is $(p_x, p_y, \theta_{\text{yaw}})$, controlled by a bounded turning rate at a fixed forward speed. 
Detailed system descriptions are deferred to the appendix. 
For the Dubins car, we study two variants: the standard model, and \emph{Dubins3D-Explore}, which augments the explorer's action space with lateral velocity to isolate the effect of control authority on active exploration. 
All systems operate under a stochastic PD controller whose output is deliberately flipped at $p_x \geq 1$, which mimics humanoid falling---where gravity overpowers stabilizing joint torques once the robot tilts beyond a critical angle---making recoverability synthesis more challenging. 
Both systems use a $2$~s horizon with $\Delta t= 0.02$~s. See Fig.~\ref{fig:toy_systems} for representative $\epsilon$-liveness set slices.

Metrics are aggregated over 5 random seeds. The ensemble uncertainty is the standard deviation across 3 member networks: $\sigma_{\text{ens}}(x) = \operatorname{std}\bigl(V_\phi(x, 0)\bigr)$. The RND uncertainty is the discrepancy $\sigma_{\text{rnd}}(x) = \lvert V_\phi(x) - V_\theta(x, 0) \rvert$ between a distillation network $V_\phi$ (supervised by $V_\theta$ on the same dataset) and the learned value at $k=0$. To ensure both estimators remain informative in OOD regions, they incorporate a strong, untrainable prior $P$, such that $V_\phi(\cdot) = \mathrm{NN}_\phi(\cdot) + P(\cdot)$. For novel seed exploration, we roll out at most 20K trajectories within each training loop, with $\epsilon_\sigma=0.1$.

\paragraph{Q1: Liveness-set coverage and design choices}
Fig.~\ref{fig:ablation} details the per-round learning curves and final metrics. Evidently, the RND estimator provides a highly efficient alternative to ensemble methods. When paired with the reach-avoid explorer, RND matches or surpasses ensemble performance while requiring a lower computational cost as it eliminates the need to train multiple member networks. It additionally provides sharper OOD detection, which effectively bounds the augmented dataset size (cumulative seeds).
 
Furthermore, reach-avoid exploration demonstrates a significant advantage over unbiased and biased random exploration, particularly in highly underactuated environments. On the Dubins car, unbiased exploration completely collapses (0.267 coverage) and biased exploration manages only a partial recovery (0.408). By contrast, the reach-avoid explorer achieves a coverage of 0.656. On the DI system, it similarly elevates coverage to 0.941, compared to $\sim$0.86 for both random baselines. More importantly, these superior results are obtained in fewer active-learning rounds, demonstrating the efficiency of the reach-avoid explorer. Driven by these margins, we adopt \textbf{RND + reach-avoid} as our default configuration for the remaining experiments.

Despite these positive outcomes, residual coverage gaps remain. Beyond minor learning errors, we attribute these gaps to the inherent underactuation of the systems rather than explorer inefficiency. 
On the DI system, the recoverable states missed by our default configuration are primarily those moving rapidly toward the target within the control-flip zone (analogous to a humanoid bouncing upward from a highly tilted configuration). 
\begin{wrapfigure}{r}{0.26\textwidth}
  \centering
  \vspace{-6pt} 
  \includegraphics[width=0.95\linewidth]{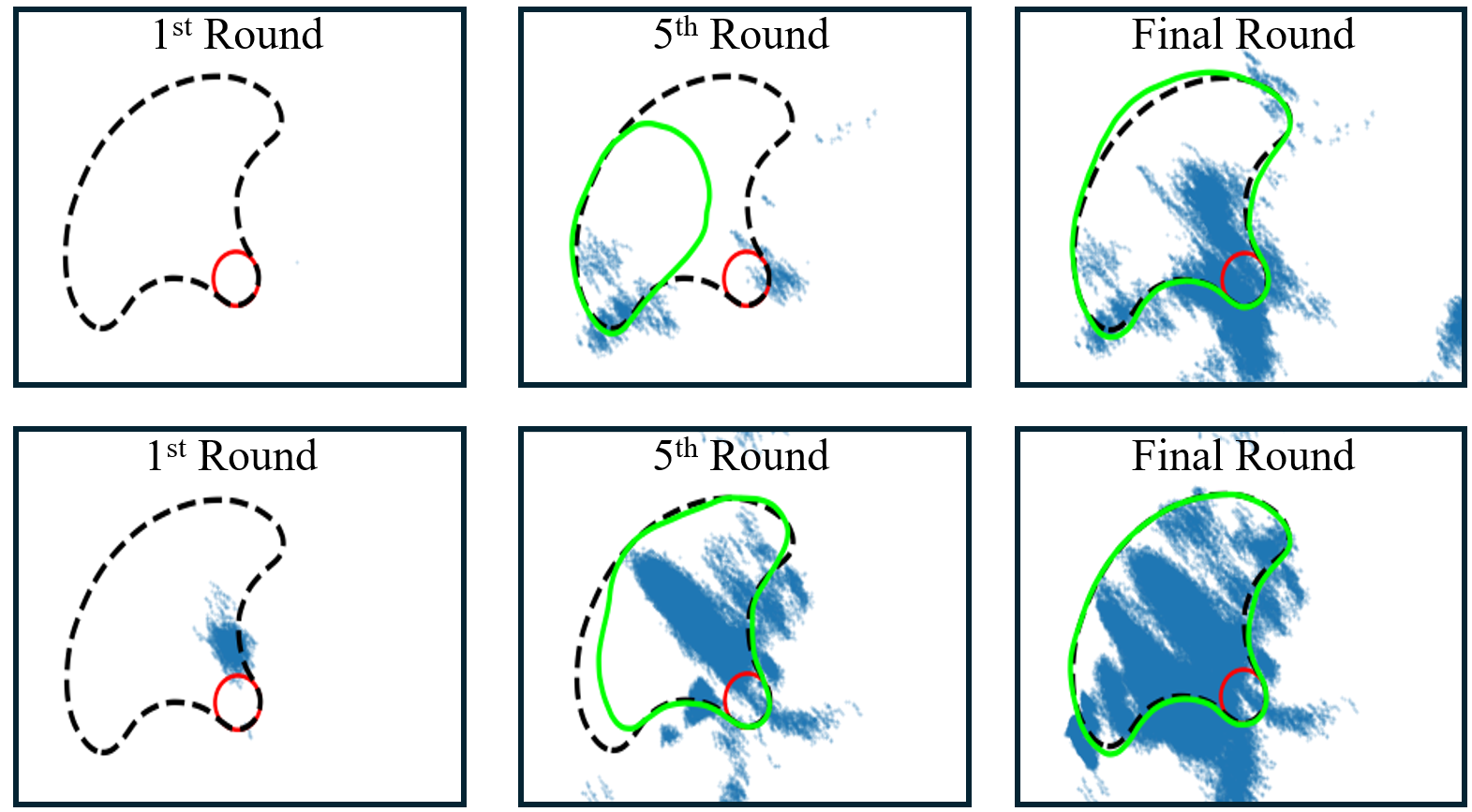}
  \vspace{-0.5em}  
  \caption{\footnotesize 3D Dubins: Data coverage evolution over rounds on an originally OOD slice ($\theta_{\text{yaw}} = -\frac{\pi}{4}$). Dashed black: GT recoverable set; Green: learned set; Blue: all datapoints with $\theta_{\text{yaw}} = -\frac{\pi}{4}\pm0.1$.}
  \vspace{-0.8em}
\label{fig:dubins_expansion}
\end{wrapfigure}
Reaching these novel states from existing data requires simultaneously increasing position $p_x$ while driving velocity $v_x$ more negative---a maneuver the closed-loop dynamics simply cannot execute.
This limitation is more pronounced in the standard Dubins car, where non-holonomic constraints render a large part of the recoverable set unreachable from $\rho_0$ without first traversing unrecoverable regions. Granting the explorer additional control authority---as in the \emph{Dubins3D-Explore} variant, which explores by controlling both turning rate and lateral velocity---closes much of this gap (Fig.~\ref{fig:dubins_expansion}). This augmented authority raises the RND + reach-avoid coverage from 0.656 to 0.921 and $F_1$ to 0.970, confirming that achievable coverage is fundamentally bounded by the explorer's action space. While the ground-truth recoverable set is not exhaustively covered, the missed states remain kinematically unreachable from $\rho_0$ without traversing unrecoverable regions or experiencing severe unmodeled deployment-time perturbations.

\paragraph{Q2: Accuracy---TD+MC vs.\ TD-only}
Since liveness values do not converge within the $2$-s horizon in the toy examples, we defer comparison against the time-agnostic TD baseline to the humanoid experiments (Sec.~\ref{subsec:res_humanoid}). Instead, we focus on validating the necessity of the MC anchoring loss by comparing against the TD-only variant. We conduct this ablation on the DI system due to its higher stochasticity.
\begin{table}[H]
\centering
\vspace{-1.0em}
\caption{\small Final-round mean results: TD-only vs TD+MC on DI.}
\label{tab:td_mc_double_integrator_rnd_reach_avoid}
\resizebox{\columnwidth}{!}{%
\begin{tabular}{lcccccc}
\toprule
Method & Precision & Recall & $F_1$ & MSE & Cum.\ seeds & Wall (min) \\
\midrule
RND + reach-avoid (TD+MC) & \textbf{0.942} & \textbf{0.955} & \textbf{0.949} & \textbf{0.006} & \textbf{53524} & 71 \\
RND + reach-avoid (TD only) & 0.885 & 0.940 & 0.912 & 0.015 & 72710 & \textbf{57} \\
\midrule
Ensemble + unbiased (TD+MC) & \textbf{0.955} & \textbf{0.909} & \textbf{0.932} & \textbf{0.010} & \textbf{86464} & 48 \\
Ensemble + unbiased (TD only) & 0.904 & 0.615 & 0.731 & 0.031 & 180467 & \textbf{35} \\
\bottomrule
\end{tabular}%
}
\vspace{-1.0em}
\end{table}
As anticipated, the TD+MC objective outperforms the TD-only baseline across all metrics (Table~\ref{tab:td_mc_double_integrator_rnd_reach_avoid}). Surprisingly, the TD-only loss yields an overall \emph{underestimation} of the liveness value (by being overly optimistic), seemingly contradicting the theoretical conservatism established by Jensen's inequality (see Remark). We attribute this to incomplete data coverage: the network, trained solely on local backups, extrapolates overly optimistic predictions into OOD states. Notably, under a weaker exploration configuration, \emph{ensemble + unbiased}, the TD-only training completely collapses into an extremely noisy and conservative approximation. In contrast, adding the MC loss provides crucial grounding to stabilize training dynamics.

\begin{figure}[H]
    \centering
    \includegraphics[width=0.95\linewidth]{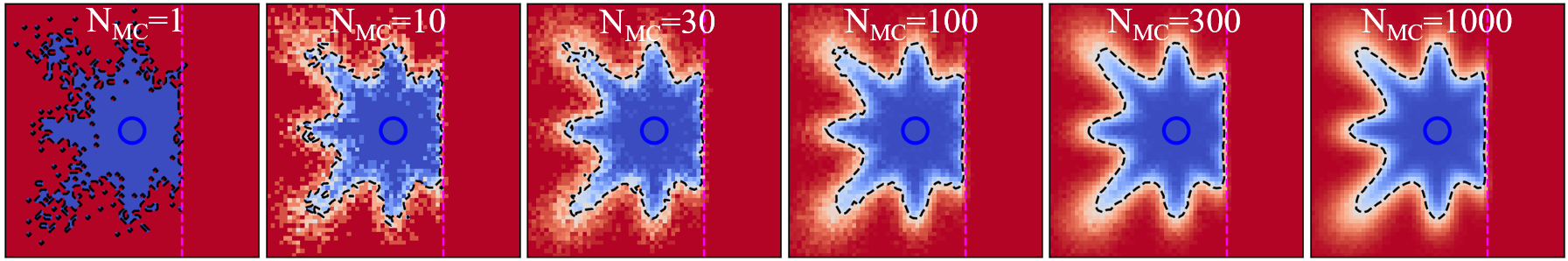}
    \caption{\footnotesize DI value heatmaps across different $N_{\text{MC}}$.}
    \label{fig:nmc_heatmaps}
    \vspace{-1.2em}
\end{figure}

\paragraph{Q3: Sensitivity to $N_{\text{MC}}$ and explorer horizon}
Table~\ref{tab:combined_sweeps} evaluates the sensitivity of our method to $N_{\text{MC}}$, the number of rollouts used to generate each MC target. Core metrics remain largely robust for $N_{\text{MC}} \geq 10$, but degrade at $N_{\text{MC}}=1$ where the target becomes a high-variance binary label (see Fig.~\ref{fig:nmc_heatmaps} for qualitative heatmaps). Here, the training efficiency scales clearly with $N_{\text{MC}}$: a larger $N_{\text{MC}}$ budget yields a cleaner grounding signal that suppresses redundant exploration. This variance reduction shrinks the augmented dataset (cumulative seeds), reduces the number of active learning rounds, and lowers total training time. Combined with the findings from Q2, this confirms that MC anchoring consistently improves stability and accuracy while introducing minimal computational overhead.
\begin{table}[H]
\centering
\caption{\small Hyperparameter sensitivity (mean $\pm$ std over seeds). Top: DI $N_{\mathrm{MC}}$ sweep (RND + reach-avoid). Bottom: 3D Dubins reach-avoid horizon sweep ($N_{\text{MC}}=1000$).}
\label{tab:combined_sweeps}
\resizebox{\columnwidth}{!}{%
\begin{tabular}{rcccccc}
\toprule
$N_{\mathrm{MC}}$ & Coverage & Precision & $F_1$ & MSE & Cum.\ seeds & Wall (min) \\
\midrule
 1 & 0.490$\pm$0.013 & 0.833$\pm$0.009 & 0.838$\pm$0.005 & 0.032$\pm$0.001 & 486626 & 80 \\
 10 & 0.830$\pm$0.006 & 0.913$\pm$0.003 & 0.920$\pm$0.001 & 0.013$\pm$0.003 & 405432 & 80 \\
 30 & 0.889$\pm$0.010 & 0.929$\pm$0.003 & 0.934$\pm$0.004 & 0.009$\pm$0.001 & 164003 & 78 \\
 100 & 0.935$\pm$0.008 & 0.918$\pm$0.008 & 0.936$\pm$0.003 & 0.008$\pm$0.000 & 60884 & 79 \\
 300 & 0.929$\pm$0.004 & 0.930$\pm$0.011 & 0.939$\pm$0.001 & 0.007$\pm$0.001 & 69518 & 79 \\
 1000 & \textbf{0.941$\pm$0.011} & \textbf{0.942$\pm$0.010} & \textbf{0.949$\pm$0.007} & \textbf{0.006$\pm$0.001} & 53524 & 71 \\
\midrule
\midrule
Horizon & Coverage & Precision & $F_1$ & MSE & Cum.\ seeds & Wall (min) \\
\midrule
1 & 0.614$\pm$0.166 & 0.966$\pm$0.012 & 0.797$\pm$0.130 & 0.085$\pm$0.070 & 29442 & 10 \\
5 & 0.604$\pm$0.151 & 0.943$\pm$0.009 & 0.803$\pm$0.089 & 0.067$\pm$0.032 & 25965 & 12 \\
10 & {\textbf{0.748$\pm$0.050}} & {\textbf{0.971$\pm$0.010}} & {\textbf{ 0.913$\pm$0.010}} & {\textbf{ 0.025$\pm$0.003}} & 27868 & 21 \\
25 & 0.740$\pm$0.137 & 0.953$\pm$0.026 & 0.878$\pm$0.092 & 0.037$\pm$0.029 & 27449 & 28 \\
50 & 0.656$\pm$0.105 & 0.897$\pm$0.023 & 0.839$\pm$0.087 & 0.064$\pm$0.036 & 30254 & 33 \\
100 & 0.682$\pm$0.059 & 0.962$\pm$0.018 & 0.871$\pm$0.040 & 0.041$\pm$0.018 & 27788 & 69 \\
\bottomrule
\end{tabular}%
}
\vspace{-1.5em}
\end{table}
Table~\ref{tab:combined_sweeps} also examines the effect of the reach-avoid explorer's training horizon. 
The ablation is conducted on the heavily underactuated Dubins system, where the horizon is expected to play a critical role. 
A moderate $10$-step horizon ($0.2$~s) optimally balances the training time against the ability to actively discover a broad range of novel states. 
Conversely, an overly myopic horizon leads to poor exploration capacity, while an overly long horizon spikes the wall time and causes the exploration to collapse to a few deeply OOD states.

\subsection{Unitree G1 Humanoid: Open-Loop Metrics}
\label{subsec:res_humanoid}
We now validate the framework's scalability by certifying a Unitree G1 balancing policy $\pi$ in a 69D state space, simulated via mjlab \cite{zakka2026mjlablightweightframeworkgpuaccelerated} at 50\,Hz. 
To specify the balancing task, we define the target function as $g(x) = \tanh(5 \cdot \text{SDF}(x))$, where $\text{SDF}(x) = \max\bigl( \|v\|, 0.25\|\omega\|, 0.005\|\dot{q}\|_{\infty}, 2.0|h - 0.77| \bigr) - 0.2$ bounds the base velocities, joint velocities $\dot{q}$, and torso height $h$ to a stationary upright posture. 
We evaluate over a 3-second liveness horizon, which extends well beyond the time typically required for $\pi$ to successfully recover balance. 
The state representation encompasses the torso's height, roll and pitch angles, linear and angular velocities, as well as full-body joint configurations and velocities. Although some inputs are unnecessary for training a robust RL policy, the recoverability monitor requires them to ensure the mapping between the physical state and the predicted value remains injective. In hardware deployment, the body-frame velocities and height are estimated using a light-weight LSTM with proprioceptive inputs. 
The initial-state distribution $\rho_0$ comprises: (1) static canonical upright postures with base linear velocities $v_x, v_y \in [-3, 3]$\,m/s, and (2) nominal walking states with forward velocity $v_x \in [-0.5, 1.5]$\,m/s, and lateral/angular velocities $v_y, \omega_z \in [-1, 1]$.

\textbf{Framework Design Choices.} Guided by the toy case studies, we configure the active learning loop with the \textbf{RND + reach-avoid} explorer operating over a moderate 0.4~s training horizon. 
We set $N_{\text{MC}} = 32$ to balance the variance-reduction benefits of MC grounding against the significantly higher computational expense of humanoid rollouts. 
Crucially, inspired by the \emph{Dubins3D-Explore} ablation, we carefully design the explorer's control authority. 
Granting the explorer full whole-body joint control in this extremely high-dimensional space is prohibitively inefficient. 
Instead, we define its action space as learned hip-joint offsets (left and right; roll, pitch, and yaw) added to the nominal walking policy, alongside a longitudinal external torso push bounded by $\pm 50$\,N, resulting in a compact 7D \emph{trainable} action space for the explorer. 
We note that this formulation scales naturally: the explorer can be easily trained to output offsets conditioned on multiple distinct nominal policies (e.g., walking, running, whole-body manipulation, etc.) simultaneously by simply augmenting its input state with a policy indicator variable. 
We train both a continuous liveness value $V_\theta$ (where $x$ is labeled recoverable iff $V_\theta(x,0) \leq 0$ for metrics computation) and a failure probability $P_{\text{fail}}(x,0)$ using binary $g(x)$ (labeled recoverable iff $P_{\text{fail}} \leq 0.1$).

\textbf{Validation and OOD Test Set Generation.} To evaluate the learned solutions, we generate a \emph{validation} set by rolling out trajectories from $\rho_0$ using held-out seeds and labeling intermediate states, alongside a challenging \emph{OOD test} set. We construct OOD states by warming up the environment with a periodic walking policy, then injecting a constant 100~N external force to a random link (pelvis, torso, or feet). This force is applied in a random horizontal direction (a body-frame forward cone for torso/pelvis, backward for feet) for exactly 1\,s. States captured immediately upon release are binned and equalized by estimated post-disturbance difficulty ($\|v\| + 0.25\,\|\omega\|$). Finally, we assign liveness value labels to these states via 100 MC rollouts of $\pi$ per state.

\begin{figure}[t]
    \centering
    \begin{minipage}{0.65\linewidth}
        \centering
        \includegraphics[width=\linewidth]{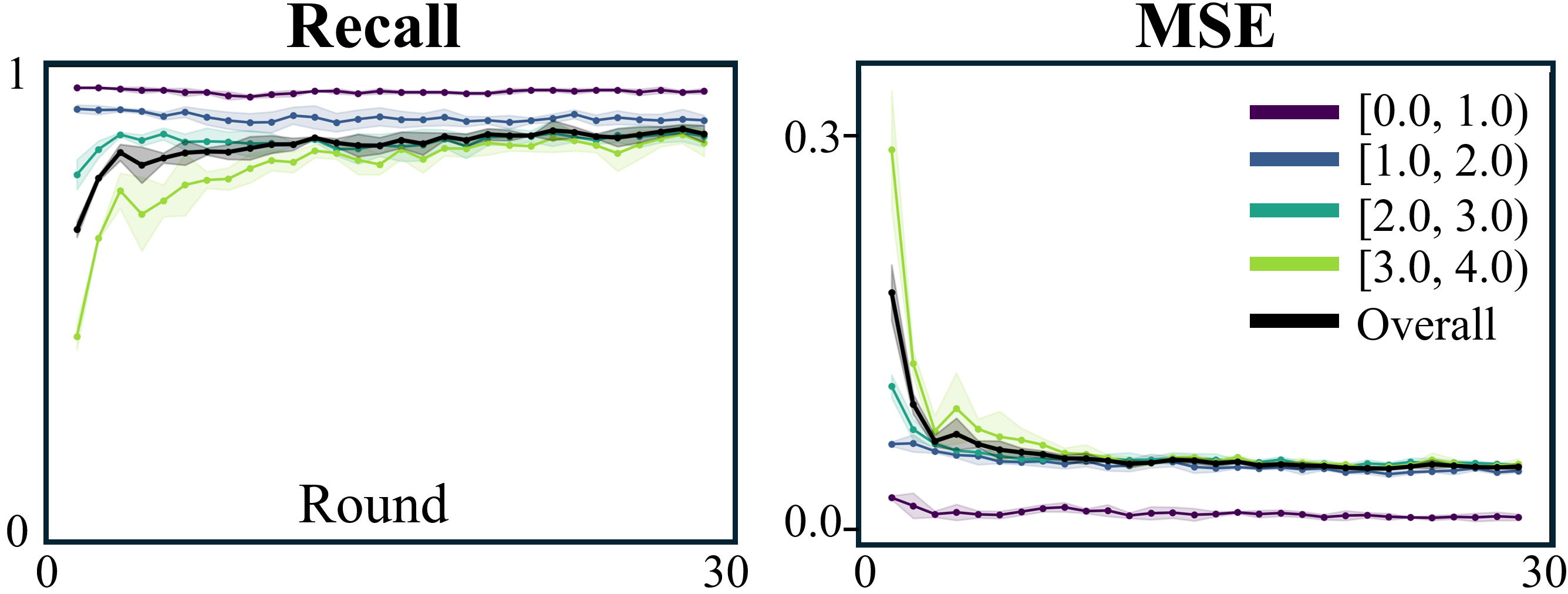}
    \end{minipage}\hfill
    \begin{minipage}{0.30\linewidth}
        \centering
        \includegraphics[width=\linewidth]{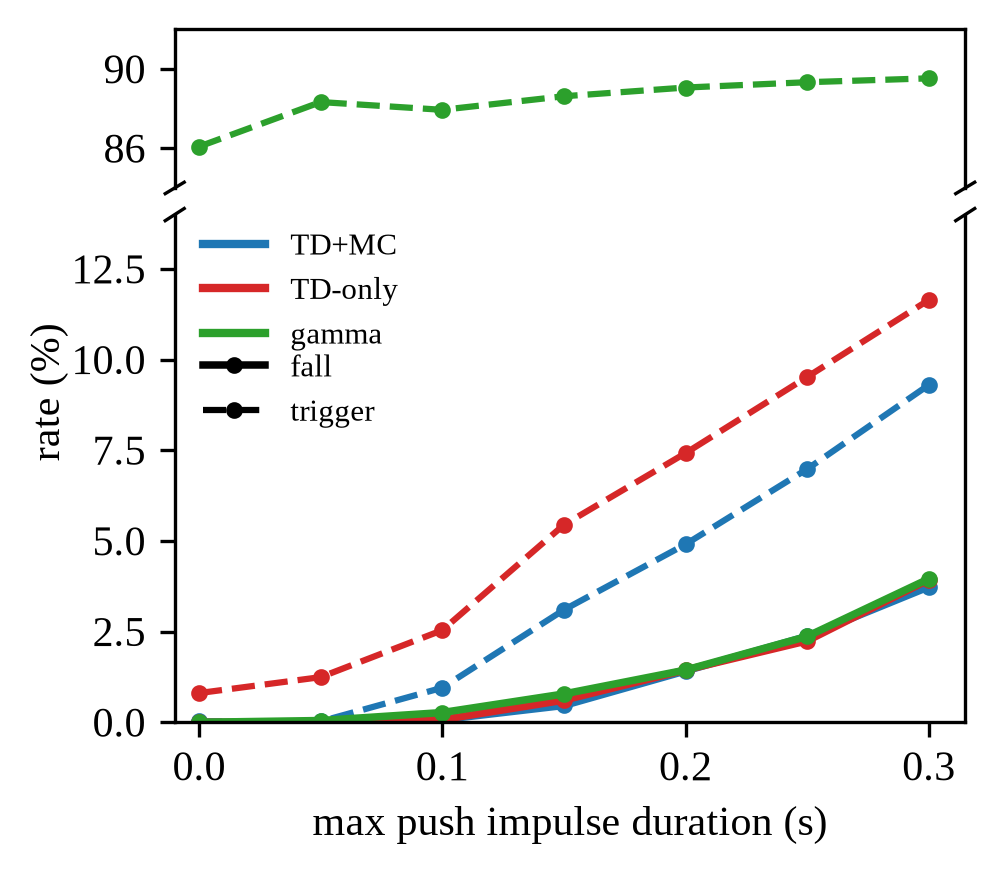}
    \end{minipage}
    \caption{\footnotesize (Left) Per-bin performance on the OOD push set (seed 0). Bins labeled by $\|v\|+0.25\|\omega\|$. (Right) Closed-loop failure/trigger rates across various maximum push impulse durations. The monitor with our TD+MC objective triggers least often while achieving the smallest failure rates.}
    \label{fig:merged_humanoid_results}
    \vspace{-1.5em}
\end{figure}

\textbf{Main Results.} After 30 active-learning rounds (Table~\ref{tab:humanoid}, $\sim$6h wall time), $V_\theta$ attains an excellent $F_1$ score of $0.957 \pm 0.001$ on the validation set and maintains a highly robust $F_1$ of $0.918 \pm 0.009$ on the aggressive OOD push dataset, yielding confident coverages of 0.944 and 0.815, respectively. The $P_{\text{fail}}$ estimator trades a marginal amount of classification accuracy ($F_1 = 0.920$ Val / $0.799$ OOD) for better interpretability. 
Notably, both $F_1$ and coverage improve almost monotonically across rounds. The most substantial gains occur in OOD recall ($V_\theta$, Fig.~\ref{fig:merged_humanoid_results}, left), surging from 0.672 to 0.893 overall, and from 0.425 to 0.871 in the most difficult $[3.0, 4.0)$ bin, as the explorer actively unearths boundary states that are unreachable by simply rolling out $\pi$ from $\rho_0$. 
A similar trend appears in the $P_{\text{fail}}$ estimator, though coverage expansion is less pronounced due to the inherently sparser gradient signal of probability learning. 
While exhibiting lower overall recall, $P_{\text{fail}}$ maintains high precision---meaning it rarely hallucinates false recoverability---which is critical for deploying a real-world monitor.
\begin{table}[t]
\centering
\caption{\small Humanoid results: mean $\pm$ std over 3 seeds.}
\label{tab:humanoid}
\resizebox{\columnwidth}{!}{%
\begin{tabular}{lccccc}
\toprule
Head & Set & Prec. & Recall & $F_1$ & Coverage \\
\midrule
\multirow{2}{*}{Value $V_\theta$}
  & Val & $0.957{\pm}0.005$ & $0.956{\pm}0.003$ & $0.957{\pm}0.001$ & $0.944{\pm}0.003$ \\
  & OOD & $0.944{\pm}0.010$ & $0.893{\pm}0.024$ & $0.918{\pm}0.009$ & $0.815{\pm}0.034$ \\
\midrule
\multirow{2}{*}{$P_{\text{fail}}$}
  & Val & $0.920{\pm}0.006$ & $0.921{\pm}0.014$ & $0.920{\pm}0.004$ & $0.920{\pm}0.014$ \\
  & OOD & $0.915{\pm}0.018$ & $0.711{\pm}0.049$ & $0.799{\pm}0.026$ & $0.698{\pm}0.047$ \\
\bottomrule
\end{tabular}%
}
\vspace{-1.5em}
\end{table}

\subsection{Runtime Monitor and Hardware Deployment}
\label{subsec:res_closedloop}
After validating the open-loop metrics, we evaluate the recoverability monitors by simulating continuous walking under domain randomization and varying velocity commands ($v_x \in [-0.5, 1.5]$\,m/s, $v_y, \omega_z \in [-0.75, 0.75]$\,rad/s). 
After a brief warm-up to establish a periodic gait, a multidirectional impulse is applied to a random body link (pelvis, torso, or foot, with tiered magnitudes of 150\,N, 250\,N, and 50\,N, respectively). 
The impulse duration is sampled uniformly from $[0, T_{\max}]$, where sweeping $T_{\max}$ from 0 to 0.30\,s scales the disturbance from benign to severe across 4,096 8-second trials per sweep. 
The learned monitor switches the system to the balancing policy if the state is predicted to be unrecoverable ($V_\theta > \epsilon_V= -0.2$) or highly uncertain ($\sigma > 0.2$). 
We evaluate the monitor both on a nominal ``healthy'' walking policy (mimicking standard deployment phases) and a ``degraded'' policy subjected to severe observation noise and sensor blackouts (mimicking rapid development phases).

\textbf{Learning Objective Comparison.} We first evaluate the monitors trained under different learning objectives on the healthy walking policy across varying push durations (Fig.~\ref{fig:merged_humanoid_results}, right). Our TD+MC monitor avoids excessive conservatism, raising zero false alarms during undisturbed walking or gentle pushes ($T_{\max} \leq 0.05$\,s). In contrast, the time-agnostic TD baseline~\cite{fisac2019bridging} degenerates completely (triggering on $>86$\% of benign states), and the TD-only variant also over-triggers on normal walking, both of which would constantly interrupt task execution in practice. Despite this improved precision, our TD+MC monitor achieves the lowest failure rates across all disturbance bands, and its trigger rate strictly bounds the actual failure rate while predicting 100\% of eventual falls in advance. Finally, even under the most severe disturbance band, our monitor provides an average lead time of 0.58\,s between a high-confidence failure prediction ($V_\theta \ge 0.9$) and torso-ground contact, securing sufficient margin to potentially activate a harm-minimizing fallback.

\begin{table}[t]
\centering
\caption{\footnotesize Closed-loop failure (fall) and trigger rates across maximum push durations for the healthy and degraded policies. Unmon.: Unmonitored; Mon.: Monitored; Rel.\ Red.: relative reduction in failure rate.}
\label{tab:closed_loop_policies}
\resizebox{\columnwidth}{!}{%
\begin{tabular}{c | cccc | cccc}
\toprule
\multirow{2}{*}{$T_{\max}$ (s)} & \multicolumn{4}{c|}{Healthy Policy (\%)} & \multicolumn{4}{c}{Degraded Policy (\%)} \\
 & Unmon.\ Fall & Mon.\ Fall & Rel.\ Red. & Trigger & Unmon.\ Fall & Mon.\ Fall & Rel.\ Red. & Trigger \\
\midrule
0.00 & 0.00 & 0.00 & -- & 0.0 & 35.01 & 10.55 & 69.9 & 93.3 \\
0.05 & 0.00 & 0.00 & -- & 0.0 & 34.23 & 10.89 & 68.2 & 93.1 \\
0.10 & 0.63 & 0.44 & 30.8 & 2.8 & 37.65 & 11.67 & 69.0 & 93.8 \\
0.15 & 2.98 & 2.59 & 13.1 & 9.2 & 41.21 & 13.28 & 67.8 & 94.4 \\
0.20 & 5.86 & 4.59 & 21.7 & 13.0 & 43.60 & 16.02 & 63.3 & 94.1 \\
0.25 & 8.59 & 7.13 & 17.0 & 16.2 & 44.04 & 18.31 & 58.4 & 94.3 \\
0.30 & 12.21 & 8.94 & 26.8 & 20.2 & 47.75 & 20.75 & 56.5 & 94.8 \\
\bottomrule
\end{tabular}%
}
\vspace{-1.8 em}
\end{table}

\textbf{Performance Across Policies.} Table~\ref{tab:closed_loop_policies} details the closed-loop failure rates across the two policies. The monitor consistently reduces failure rates across all disturbance tiers, especially on the degraded walker. This confirms both the accuracy of the liveness value and the improved robustness of the fallback balancing policy. 
Aside from minor learning errors, the non-zero residual monitored failure rates stem from the highly stochastic nature of our simulation environment (aggressive domain randomization and observation noise). 
However, a physical robot represents only a single instantiation of these randomized parameters. 
Assuming the actual hardware dynamics fall well within the domain randomization bounds and physical sensor noise is more lenient, the empirical failure probability during real-world deployment is expected to be substantially lower. 
Finally, we qualitatively demonstrate the zero-shot performance on the G1 hardware in Fig.~1 and the supplementary video.

%% file: conclusion.tex
In this work, we presented a framework combining a TD+MC learning objective with uncertainty-guided reach-avoid exploration to learn policy recoverability under stochastic dynamics. Our approach successfully scales to a 29-DoF humanoid, yielding an accurate and efficient runtime monitor.
To our knowledge, this is the first effort to comprehensively map the liveness-set envelope for high-dimensional, contact-rich systems. Despite achieving promising outcomes, the framework currently acts reactively, rather than proactively, to OOD inputs. Furthermore, while empirically accurate, the learned solutions remain uncalibrated. Lastly, the maximum achievable coverage is fundamentally bounded by the choice of the explorer's action space.

Looking forward, although our framework is theoretically amenable to any contact-rich system, its efficacy on complex dexterous manipulation, long-horizon or extremely dynamic tasks requires further investigation. In addition, extending it to general reach-avoid settings, where the robot must achieve goals while obeying safety specifications, will significantly broaden its practical impact. Ultimately, leveraging these learned reachability values to improve the quality and efficiency of policy synthesis stands as a natural next step.

%% file: appendix.tex
\section*{APPENDIX}
\label{sec:appendix}
\textbf{Planar Double Integrator}
Let $\mathbf{p} = (p_x,p_y)$ and $\mathbf{v} = (v_x, v_y)$. The state is $\mathbf{x} = (\mathbf{p}, \mathbf{v}) \in \mathcal{X} = [-2.5, 2.5]^2 \times [-2, 2]^2$, with initial states $\rho_0$ drawn uniformly from $[-2, -1]^2 \times [-2, 2]^2$. The dynamics are $\dot{\mathbf{p}} = \mathbf{v} + \mathbf{d}$ and $\dot{\mathbf{v}} = \mathbf{a}$, with a target set $\mathcal{G} = \{\mathbf{x} : \|\mathbf{p}\| \leq 0.25\,\text{m}\}$. The nominal control is a noisy P-controller ($K_p = 2.0$) with a saturation bound along each axis: $\mathbf{a} = \mathrm{clip}(-K_p \mathbf{p} + \epsilon_a, -a_{\max}, a_{\max})$, where $a_{\max} = 1.0$ and $\epsilon_a \sim \mathcal{N}(0, 0.5^2 I)$. As mentioned, the controller is flipped to steer \emph{away} from the target for states where $p_x \geq 1.0$. The disturbance is $\mathbf{d} \sim \mathcal{N}(\beta \hat{\mathbf{p}}, \sigma_d(\mathbf{x})^2 I)$, featuring an outward radial bias ($\beta = 0.2$, $\hat{\mathbf{p}} = \mathbf{p}/\|\mathbf{p}\|$) and distance-scaled noise $\sigma_d(\mathbf{x}) = 0.1 + 0.75\|\mathbf{p}\|$.

\textbf{Dubins Car}
A unicycle benchmark with state $\mathbf{x} = (p_x, p_y, \theta) \in \mathcal{X} = [-2.5, 2.5]^2 \times [-\pi, \pi]$ moving at a constant speed $v = 1.0$\,m/s, with initial states $\rho_0$ drawn uniformly from $[-2, -1]^2 \times [-\pi, \pi]$. The dynamics are $\dot{p}_x  = v\cos\theta + d_x$, $\dot{p}_y = v\sin\theta + d_y$, and $\dot{\theta} = \omega$, sharing the same target $\mathcal{G}$ as the double integrator. The nominal policy is a P-controller ($K_p = 2.0$) steering the system to head toward the origin: $\omega = \mathrm{clip}(K_p\,\mathrm{wrap}(\theta_{\text{des}} - \theta) + \epsilon_\omega, -\omega_{\max}, \omega_{\max})$, with $\omega_{\max} = 1.0$, $\epsilon_\omega \sim \mathcal{N}(0, 1.0^2)$, and $\theta_{\text{des}} = \operatorname{atan2}(-p_y, -p_x)$. Disturbances are biased: $d_x \sim \mathcal{N}(-0.1, 0.5^2)$ and $d_y \sim \mathcal{N}(0.1, 0.5^2)$. The \textbf{Dubins3D-Explore} variant augments the exploration control to $u = (\omega, v_{\text{lat}})$ with $|v_{\text{lat}}| \leq 1.0$\,m/s, adding a lateral push perpendicular to the heading: $\dot{p}_x = v\cos\theta - v_{\text{lat}}\sin\theta$ and $\dot{p}_y = v\sin\theta + v_{\text{lat}}\cos\theta$.